\documentclass{article}
\usepackage{spconf,amsmath,graphicx}

\usepackage{enumitem}
\setlist{nosep, leftmargin=14pt}

\usepackage{mwe} 
\usepackage[normalem]{ulem}
\usepackage{booktabs}
\usepackage{hyperref}
\hypersetup{
    colorlinks=true,
    linkcolor=blue,
    filecolor=blue,
    urlcolor=blue,
}
\usepackage{multirow}

\title{Prompt-Free Lightweight SAM Adaptation for Histopathology Nuclei Segmentation with Strong Cross-Dataset Generalization}
\name{Muhammad Hassan Maqsood$^{1}$,Yanming Zhu$^{1}$, Alfred Lam$^{2}$,Getamesay Dagnaw$^{1}$, Xuefei Yin$^{1}$, Alan Liew$^{1}$
}
\address{%
$^{1}$ School of Information and Communication Technology, Griffith University, QLD, Australia \\
$^{2}$ School of MDP -- Clinical Medicine, Griffith University, QLD, Australia
}
\begin{document}
%
\maketitle

\begin{abstract}
Histopathology nuclei segmentation is crucial for quantitative tissue analysis and cancer diagnosis. Although existing segmentation methods have achieved strong performance, they are often computationally heavy and show limited generalization across datasets, which constrains their practical deployment. Recent SAM-based approaches have shown great potential in general and medical imaging, but typically rely on prompt guidance or complex decoders, making them less suitable for histopathology images with dense nuclei and heterogeneous appearances. We propose a prompt-free and lightweight SAM adaptation that leverages multi-level encoder features and residual decoding for accurate and efficient nuclei segmentation. The framework fine-tunes only LoRA modules within the frozen SAM encoder, requiring just 4.1M trainable parameters. Experiments on three benchmark datasets TNBC, MoNuSeg, and PanNuke demonstrate state-of-the-art performance and strong cross-dataset generalization, highlighting the effectiveness and practicality of the proposed framework for histopathology applications.
\end{abstract}


\section{Introduction}
Nuclei segmentation in histopathology is a fundamental step for quantitative tissue analysis and cancer diagnosis. Accurate delineation of nuclei is crucial for downstream tasks such as tumor grading, biomarker quantification, and prognosis prediction. Over the past decade, deep learning-based methods, such as U-Net \cite{ronneberger2015u} and its numerous variants \cite{oktay2018attention, cao2022swin, wang2022uctransnet}, have achieved remarkable progress. However, despite strong in-domain performance, many of these models are computationally heavy and exhibit limited generalization when applied to images acquired under different staining protocols, tissue types, or imaging conditions. These factors restrict their scalability and practicality in real-world clinical settings.

To enhance segmentation accuracy and representation capacity, much work has focused on architectural improvements to U-Net. Examples include the incorporation of attention mechanisms \cite{oktay2018attention}, refined skip connections \cite{lachinov2021projective}, recurrent modules \cite{li2019cr}, Gaussian attention \cite{roy2024gru}, and spatial–channel attention \cite{fu2024tsca}. More recently, Transformer-based models such as UCTransNet \cite{wang2022uctransnet}, Swin-UNet \cite{cao2022swin}, and MedT \cite{valanarasu2021medical} have improved the modeling of long-range dependencies and achieved remarkable performance. Nevertheless, these architectures often remain large and are sensitive to domain shifts, leading to performance degradation in cross-dataset evaluations.

In parallel, the emergence of vision foundation models has opened new opportunities for more generalizable medical image analysis. The Segment Anything Model (SAM) \cite{kirillov2023segment} demonstrates strong zero-shot capabilities on natural images and has inspired adaptations for medical tasks. For example, Trans-SAM \cite{wu2025trans} employs perceptual and convolutional adapters, KnowSAM \cite{huang2025learnable} leverages semi-supervised prompting and knowledge distillation, and Zhang et al. \cite{zhang2023customized} explore LoRA-based fine-tuning \cite{hu2021lora}. While promising, these methods typically depend on prompt guidance or complex decoders and have not been thoroughly evaluated on histopathology nuclei segmentation, where dense nuclei, fine textures, and stain variability present unique challenges.

To address these gaps, we propose a prompt-free and lightweight adaptation of SAM tailored for histopathology nuclei segmentation. Our framework fine-tunes only LoRA modules within the frozen SAM encoder, enabling efficient training with only 4.1M parameters. Multi-level encoder features are decoded through lightweight residual blocks to enhance boundary precision while maintaining compactness. Extensive experiments on three benchmark datasets (TNBC, MoNuSeg, and PanNuke) demonstrate state-of-the-art performance and strong cross-dataset generalization, highlighting the effectiveness and practicality of the proposed method.

\section{Method}

\begin{figure*}[!t]
    \centering
    \includegraphics[width=0.95\textwidth]{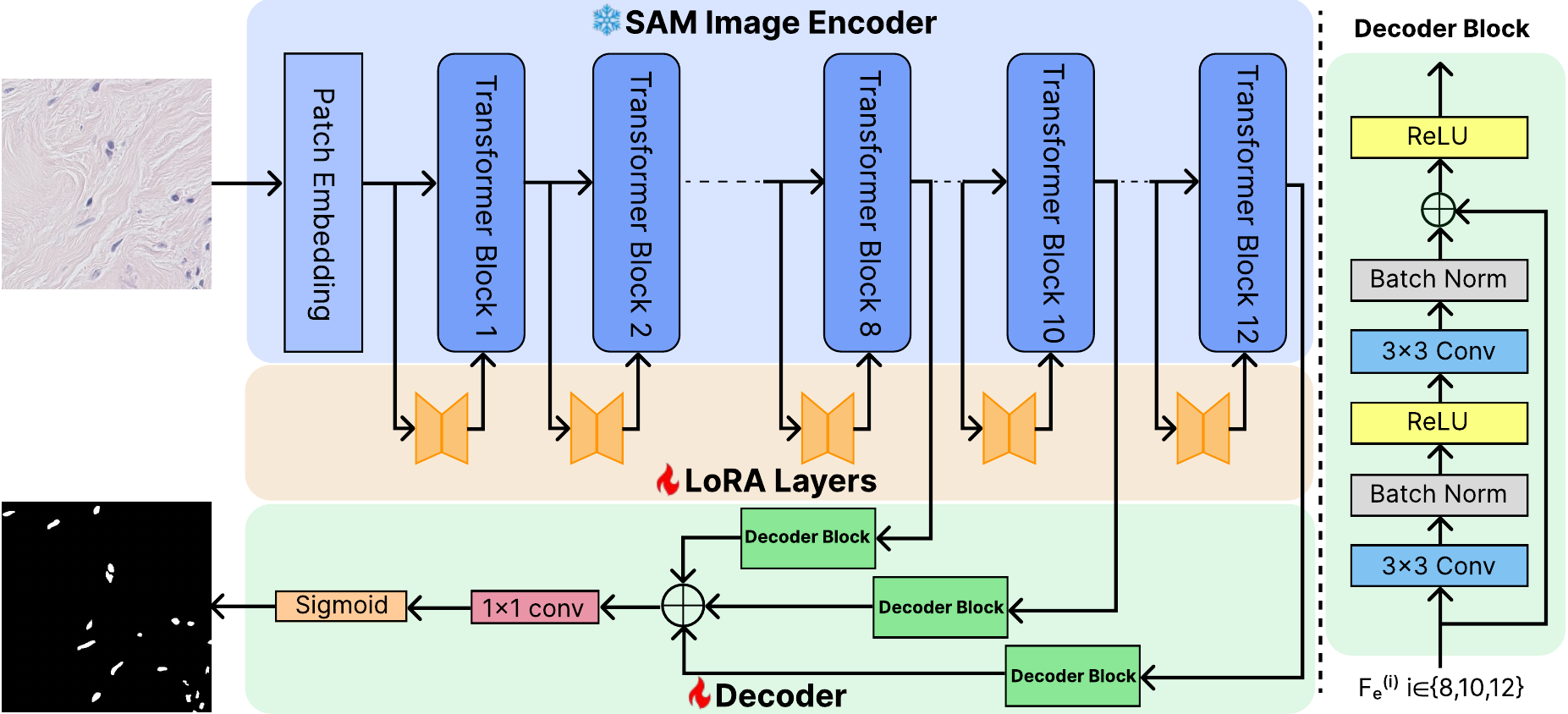}
    \caption{Overview of the proposed framework. A frozen SAM image encoder is adapted with LoRA modules inserted into its attention layers for parameter-efficient fine-tuning. Multi-level features from selected Transformer blocks are processed by lightweight residual decoder blocks and then fused to produce the final segmentation mask.}
    \label{fig:framework}
\end{figure*}

The overall framework is illustrated in Fig.\ref{fig:framework}. It consists of a frozen SAM image encoder, LoRA layers, and a residual decoder. LoRA modules are inserted into the encoder’s attention layers, and only these modules are optimized during training, keeping the number of trainable parameters very low. Intermediate feature maps are extracted from multiple Transformer blocks to capture semantic information at different abstraction levels. These features are decoded by lightweight residual blocks inspired by ResNet \cite{he2016deep}, concatenated, and passed through a $1 \times 1$ convolution followed by a sigmoid activation to produce the final segmentation mask.

\subsection{SAM Image Encoder for Feature Extraction}
We adopt the SAM image encoder as the backbone for feature extraction because its pretrained representations provide strong contextual information and generalization across diverse image domains. Such generalization is particularly valuable in histopathology image computing, where tissue and staining variations commonly degrade model robustness.

Instead of full fine-tuning, we apply LoRA \cite{hu2021lora} to efficiently adapt SAM (ViT-B) to histopathology nuclei segmentation. LoRA modules are inserted into the attention layers of all Transformer blocks within the SAM encoder, while the original encoder weights remain frozen. This design significantly reduces trainable parameters (4.1M) and training cost, while retaining SAM’s pretrained knowledge, which is crucial for robust adaptation with limited domain-specific data.

To effectively leverage both local and global information, we extract intermediate feature maps from Blocks 8, 10, and 12 of the encoder. Mid-level layers capture fine textures and boundaries, while deeper layers encode broader contextual structures. This combination provides a balanced representation that supports both precise boundary localization and contextual consistency \cite{ghiasi2022vision}. The choice of these layers is guided by their semantic diversity and computational efficiency, and its effectiveness is further validated in our ablation studies.

\subsection{Residual Decoder Block}
Most SAM-based methods rely on the original heavy, prompt-dependent decoder. We design a lightweight residual decoder to fuse multi-level encoder abstractions. Inspired by ResNet [15], the decoder refines spatial details while preserving context. Each block consists of two $3\times3$ convolutions with batch normalization and ReLU, followed by an identity skip connection. This design facilitates stable gradient flow and sharpens boundaries while keeping the model compact.

\subsection{Feature Fusion and Mask Prediction}
To integrate complementary information from different abstraction levels, decoded feature maps are concatenated to form a unified representation. This simple fusion avoids the complexity and computational overhead of multi-stage fusion schemes while maintaining a good balance between fine-grained detail and semantic context. The fused features are passed through a $1 \times 1$ convolution followed by a sigmoid activation to produce the final segmentation mask.

We further stabilize training by initializing the bias of the final convolution using the empirical nuclei foreground ratio:
\[b_0 = \log\left(\frac{\pi}{1-\pi}\right)\]
where $\pi$ denotes the average nuclei pixel ratio. This initialization aligns the initial predictions with the expected class prior, accelerating convergence and improving generalization.

\subsection{Objective Function}
To address class imbalance and emphasize difficult regions, we adopt the focal Tversky loss \cite{abraham2019novel}, a well-established loss function widely used in medical image segmentation. The weighting parameters $\alpha$ and $\beta$ control the trade-off between false positives and false negatives, while $\gamma$ focuses the optimization on hard-to-segment nuclei. Following prior work \cite{abraham2019novel} and minor tuning on the validation set, we set $\alpha = 0.6$, $\beta = 0.4$, $\gamma = 2.5$, and $\epsilon = 10^{-6}$. This configuration provides stable optimization and encourages accurate segmentation of challenging nuclear regions.

\section{Experiments and Results}
\subsection{Datasets and Implementation Details}
We evaluate our framework on three widely used histopathology nuclei segmentation benchmarks: Triple Negative Breast Cancer (TNBC) \cite{naylor2018segmentation}, Multi-Organ Nuclei Segmentation (MoNuSeg) \cite{kumar2019multi}, and PanNuke \cite{gamper2019pannuke}. TNBC and MoNuSeg images are randomly split into training, validation, and test sets (80/10/10\%). PanNuke, covering 19 tissue types, is evaluated via three experiments using the official folds, rotating each for training, validation, and testing; we report the overall mean. All images are cropped into 512×512 patches without augmentation. Training uses the AdamW optimizer \cite{loshchilov2017decoupled} (initial LR $1\times10^{-4}$) with a ReduceLROnPlateau scheduler for 100 epochs. Performance is measured using Dice and Intersection over Union (IoU) metrics averaged over the test sets.

We use Dice coefficient and Intersection over Union (IoU) as evaluation metrics. Dice reflects pixel-wise segmentation accuracy, while IoU penalizes over- and under-segmentation, providing complementary perspectives on performance. All results are averaged over test sets, with PanNuke results reported as mean across the three folds.

\subsection{Results and Discussion}
\textbf{Benchmark Performance.} We compare our method with representative CNN-, Transformer-, and SAM-based baselines on TNBC, MoNuSeg, and PanNuke datasets. The compared methods include U-Net \cite{ronneberger2015u}, Swin-Unet \cite{cao2022swin}, UCTransNet \cite{wang2022uctransnet}, TSCA-Net \cite{fu2024tsca}, the original SAM \cite{kirillov2023segment}, its LoRA-adapted variant \cite{hu2021lora}, and the latest SAM adaptation for medical image segmentation, Trans-SAM \cite{wu2025trans}. As shown in Table \ref{tab:supervised}, our method achieves the highest performance scores across all datasets, outperforming both CNN- and Transformer-based methods as well as SAM baselines. Notably, despite having only 4.1M trainable parameters, our model surpasses larger models such as UCTransNet and TSCA-Net, demonstrating strong parameter efficiency. These results confirm that LoRA-based SAM adaptation with residual decoding and multi-level feature fusion jointly enhances segmentation accuracy and robustness.

\begin{table}[t]
\centering
\caption{Quantitative comparison of segmentation performance on the TNBC, MoNuSeg, and PanNuke datasets. \textbf{Bold} indicates the best results. \underline{Underline} denotes the second best.}
\resizebox{\columnwidth}{!}{
\setlength{\tabcolsep}{1pt}
\begin{tabular}{l ccc ccc ccc c}
\toprule[1pt]
\multirow{2}{*}{\textbf{Model}} & \multicolumn{2}{c}{\textbf{TNBC}} && \multicolumn{2}{c}{\textbf{MoNuSeg}} && \multicolumn{2}{c}{\textbf{PanNuke}} && \textbf{Params} \\
\cline{2-3}\cline{5-6}\cline{8-9}
 & Dice (\%) & IoU (\%) && Dice (\%) & IoU (\%) && Dice (\%) & IoU (\%) && (M) \\
\midrule[1pt]
UNet \cite{ronneberger2015u} & 67.84 & 52.46 && 68.99 & 53.20 && 71.41 & 58.40 && 31 \\
Swin-Unet \cite{cao2022swin} & 75.89 & 61.67 && 78.25 & 64.61 && 77.63 & 65.09 && 49 \\
UCTransNet \cite{wang2022uctransnet} & 79.13 & 65.83 && 79.08 & 65.50 && \underline{79.71} & \underline{69.40} && 66 \\
TSCA-Net \cite{fu2024tsca} & \underline{81.90} & \underline{69.75} && \underline{80.23} & \underline{67.13} && -- & -- && $\sim25$ \\
\midrule
SAM \cite{kirillov2023segment} & 62.56 & 49.87 && 76.60 & 62.30 && 66.03 & 52.76 && 93 \\
SAM + LoRA & 62.36 & 49.54 && 76.47 & 62.14 && 65.42 & 51.96 && 0.19 \\
Trans-SAM \cite{wu2025trans} & 62.09 & 45.68 && 56.92 & 40.99 && 68.22 & 53.06 && 96 \\
\midrule
Ours & \textbf{82.90} & \textbf{70.90} && \textbf{82.85} & \textbf{70.85} && \textbf{83.60} & \textbf{73.26} && 4.1 \\
\bottomrule[1pt]
\end{tabular}%
\label{tab:supervised}
}
\end{table}

\renewcommand{\arraystretch}{1.1}
\begin{table}[htb]
\centering
\setlength{\tabcolsep}{1pt}
\caption{Cross-dataset generalization of nuclei segmentation models, reported as Dice/IoU percentages. \textbf{Bold} indicates the best result and \underline{underline} the second best.}
\resizebox{\columnwidth}{!}
{
\begin{tabular}{l ccc ccc cc}
\toprule[1pt]
\multirow{2}{*}{\textbf{Model}} & \multicolumn{2}{c}{\textbf{TNBC} $\rightarrow$} && \multicolumn{2}{c}{\textbf{MoNuSeg} $\rightarrow$} && \multicolumn{2}{c}{\textbf{PanNuke} $\rightarrow$} \\
\cline{2-3}\cline{5-6}\cline{8-9}
 & \textbf{MoNuSeg} & \textbf{PanNuke} && \textbf{TNBC} & \textbf{PanNuke} && \textbf{TNBC} & \textbf{MoNuSeg} \\
\midrule[1pt]
UNet \cite{ronneberger2015u} & 66.0/49.91 & \underline{68.03/54.36} && 16.20/10.31 & \underline{66.49/53.17} && 12.39/7.28 & 51.43/34.90 \\
Swin-Unet \cite{cao2022swin} & 68.81/52.88 & 63.63/49.15 && 41.92/30.52 & 56.63/41.22 && 39.40/27.84 & 74.20/59.14 \\
UCTransNet \cite{wang2022uctransnet} & 61.70/45.42 & 58.73/44.16 && \underline{68.80/53.58} & 51.70/36.73 && 37.20/28.53 & 36.38/24.36 \\
TSCA-Net \cite{fu2024tsca} & \textbf{89.01}/\underline{65.97} & -- / -- && 46.76/41.36 & -- / -- && -- / -- & -- / -- \\
\midrule
SAM \cite{kirillov2023segment} & 78.95/65.43 & 26.84/16.51 && 51.28/39.51 & 54.35/40.34 && \underline{60.45/44.89} & \underline{76.16/61.81} \\
SAM + LoRA  & 78.80/65.43 & 52.16/38.64 && 52.09/39.92 & 54.15/40.51 && 55.17/40.15 & 76.02/61.53 \\
Trans-SAM \cite{wu2025trans}  & 69.81/53.76 & \textbf{71.57/58.68} && 17.94/10.61 & 49.25/36.73 && 48.29/33.68 & 56.31/41.42 \\
\midrule
Ours & \underline{80.19}/\textbf{67.06} & 60.56/47.33 && \textbf{79.45/66.18} & \textbf{70.73/57.83} && \textbf{68.81/54.57} & \textbf{80.65/67.66} \\
\bottomrule[1pt]
\end{tabular}%
\label{tab:crossdomain}
}
\end{table}

\begin{figure*}[t]
    \centering
    \includegraphics[width=\textwidth]{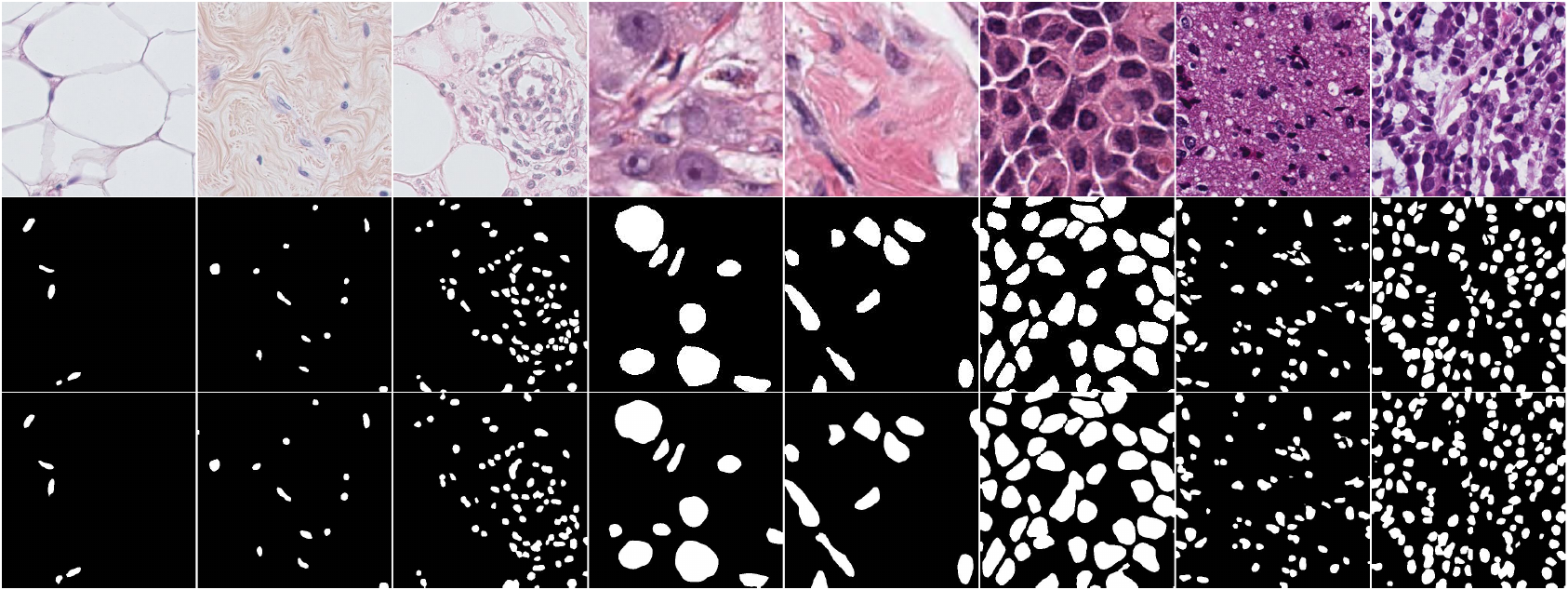}
    \caption{Qualitative segmentation results on TNBC (Col 1-3), PanNuke (Col 4-5), and MoNuSeg (Col 6-8) datasets. Input images (top), ground truth annotations (middle), and predicted masks (bottom) are shown.}
    \label{fig:qualitative}
\end{figure*}

\textbf{Cross-Dataset Generalization.} To assess generalization capability, each model is trained on one dataset and directly tested on the remaining two without fine-tuning. Table \ref{tab:crossdomain} reports the Dice and IoU scores across all transfer settings among TNBC, MoNuSeg, and PanNuke. Our method achieves the best or second-best performance in nearly all transfer directions, consistently outperforming CNN-, Transformer-, and SAM-based baselines. The performance gain is particularly pronounced when transferring from MoNuSeg and PanNuke to unseen domains, indicating strong adaptability to large variations in tissue type and staining conditions. These results provide clear evidence supporting our claim that the proposed framework significantly enhances the transferability and robustness of SAM features. Unlike existing baselines, our method maintains both high accuracy and stability across diverse domains, demonstrating its potential for deployment in real-world clinical settings where domain shifts are inevitable.
\textbf{Qualitative Results.} Representative qualitative results from TNBC, MoNuSeg, and PanNuke datasets are presented in Fig. \ref{fig:qualitative}. The proposed method accurately delineates nuclei boundaries under diverse tissue types, staining conditions, and nuclei densities. It effectively separates tightly clustered nuclei while maintaining structural consistency in sparse areas, and its predicted masks exhibit strong agreement with the ground truth, demonstrating both high precision and strong generalization to visually diverse histopathology samples.

\subsection{Ablation Studies}
To evaluate the contribution of each component, we perform ablation experiments by selectively removing key modules, as summarized in Table \ref{tab:ablations_highlevel}. Specifically, we assess the impact of (1) LoRA-based encoder adaptation, (2) bias-prior initialization, and (3) multi-level feature integration.

\textbf{Effect of LoRA adaptation.} Removing LoRA modules (“No LoRA”) leads to a consistent performance drop across all datasets, confirming that lightweight encoder adaptation is essential for transferring SAM representations to histopathology nuclei segmentation.

\textbf{Effect of Bias-prior initialization.} Excluding the bias prior (“No bias prior”) causes small but consistent declines, demonstrating that initializing the final convolution with the empirical prior stabilizes early training and improves convergence, particularly on smaller datasets such as TNBC.

\textbf{Effect of multi-level Feature integration.} We compare single-level (“Single-Level (12)”) and dual-level (“Dual-Level (10,12)”) feature fusion with the full multi-level fusion (“Ours”). The single-level variant performs slightly better on MoNuSeg but degrades on TNBC and PanNuke with denser nuclei. Dual-level fusion improves moderately yet remains below the full model, which fuses Blocks 8, 10, and 12 for balanced texture and context.

Overall, the full framework achieves the best or near-best results across all datasets, confirming the complementary benefits of each component.

\setlength{\tabcolsep}{2pt}
\begin{table}[t]
\centering
\caption{Ablation studies on TNBC, MoNuSeg, and PanNuke datasets. \textbf{Bold} indicates the best result and \underline{underline} the second best.}
\resizebox{\columnwidth}{!}{%
\begin{tabular}{l ccc ccc cc}
\toprule[1pt]
\multirow{2}{*}{\textbf{Variant}} & \multicolumn{2}{c}{\textbf{TNBC}} && \multicolumn{2}{c}{\textbf{MoNuSeg}} && \multicolumn{2}{c}{\textbf{PanNuke}} \\
\cline{2-3}\cline{5-6}\cline{8-9}
 & Dice (\%) & IoU (\%) && Dice (\%) & IoU (\%) && Dice (\%) & IoU (\%) \\
\midrule[1pt]
No LoRA & 81.60 & 69.22 && 82.74 & 70.65 && 82.11 & 71.10 \\
No bias prior & 82.14 & 70.10 && 82.74 & 70.68 && \underline{83.36} & \underline{73.03} \\
Single-Level (12) & 79.45 & 66.10 && \textbf{83.28} & \textbf{71.46} && 82.97 & 72.40 \\
Dual-Level (10,12) & \underline{82.42} & \underline{70.58} && 82.78 & 70.73 && 83.14 & 72.69 \\
\midrule
\textbf{Ours (full)} & \textbf{82.90} & \textbf{70.90} && \underline{82.85} & \underline{70.85} && \textbf{83.60} & \textbf{73.26} \\
\bottomrule[1pt]
\end{tabular}%
}
\label{tab:ablations_highlevel}
\end{table}

\section{Conclusion}
We presented a lightweight, prompt-free SAM adaptation for histopathology nuclei segmentation. The proposed framework fine-tunes only LoRA modules within the frozen SAM encoder and introduces residual decoding with multi-level feature fusion and bias-prior initialization. Despite having only 4.1M trainable parameters, it achieves state-of-the-art performance and strong cross-dataset generalization on TNBC, MoNuSeg, and PanNuke. These results demonstrate that efficient SAM adaptation combined with task-specific decoding can deliver accurate and robust segmentation across diverse histopathology domains.

\bibliographystyle{IEEEbib}
\bibliography{strings}

\end{document}